\documentclass{bmvc2k}
\usepackage{amsmath}
\usepackage{mathtools}
\usepackage[outercaption]{sidecap} 
\usepackage{wrapfig,lipsum,booktabs}

\usepackage{times}
\usepackage{epsfig}
\usepackage{graphicx}
\usepackage{amsmath}
\usepackage{amssymb}
\usepackage{booktabs}
\usepackage{hyperref}
\usepackage{pifont}
\usepackage[accsupp]{axessibility}  
\newcommand{\xmark}{\ding{55}}
\usepackage{multirow}
\usepackage{color}
\usepackage{graphicx,multirow}

\title{SA2-Net: Scale-aware Attention Network for Microscopic Image Segmentation}

\addauthor{Mustansar Fiaz}{mustansar.fiaz@mbzuai.ac.ae}{1}
\addauthor{Moein Heidari}{moeinheidari7829@gmail.com}{2}
\addauthor{Rao Muhammad Anwer}{rao.anwer@mbzuai.ac.ae}{1}
\addauthor{Hisham Cholakkal}{hisham.cholakkal@mbzuai.ac.ae}{1}
\addinstitution{
Mohamed bin Zayed University of Artificial Intelligence \\
Abu Dhabi, UAE\\
}
\addinstitution{
 Iran University of Science and Technology,
Iran
}

\runninghead{Student, Prof, Collaborator}{BMVC Author Guidelines}

\begin{document}

\maketitle

\begin{abstract}
Microscopic image segmentation is a challenging task, wherein the objective is to assign semantic labels to each pixel in a given microscopic image. While convolutional neural networks (CNNs) form the foundation of many existing frameworks, they often struggle to explicitly capture long-range dependencies. Although transformers were initially devised to address this issue using self-attention, it has been proven that both local and global features are crucial for addressing diverse challenges in microscopic images, including variations in shape, size, appearance, and target region density. In this paper, we introduce SA2-Net, an attention-guided method that leverages multi-scale feature learning to effectively handle diverse structures within microscopic images. Specifically, we propose a scale-aware attention (SA2) module designed to capture inherent variations in scales and shapes of microscopic regions, such as cells, for accurate segmentation. This module incorporates local attention at each level of multi-stage features, as well as global attention across multiple resolutions. Furthermore, we address the issue of blurred region boundaries (e.g., cell boundaries) by introducing a novel upsampling strategy called the Adaptive Up-Attention (AuA) module. This module enhances discriminative ability for improved localization of microscopic regions using an explicit attention mechanism. 
Extensive experiments on five challenging datasets  demonstrate the benefits of our SA2-Net model.  Our source code is publicly available at \url{https://github.com/mustansarfiaz/SA2-Net}. 

\end{abstract}



\section{Introduction}
\label{sec:intro}
\begin{wrapfigure}{r}{0.45\textwidth}
\centering
    \includegraphics[width=0.45\textwidth]{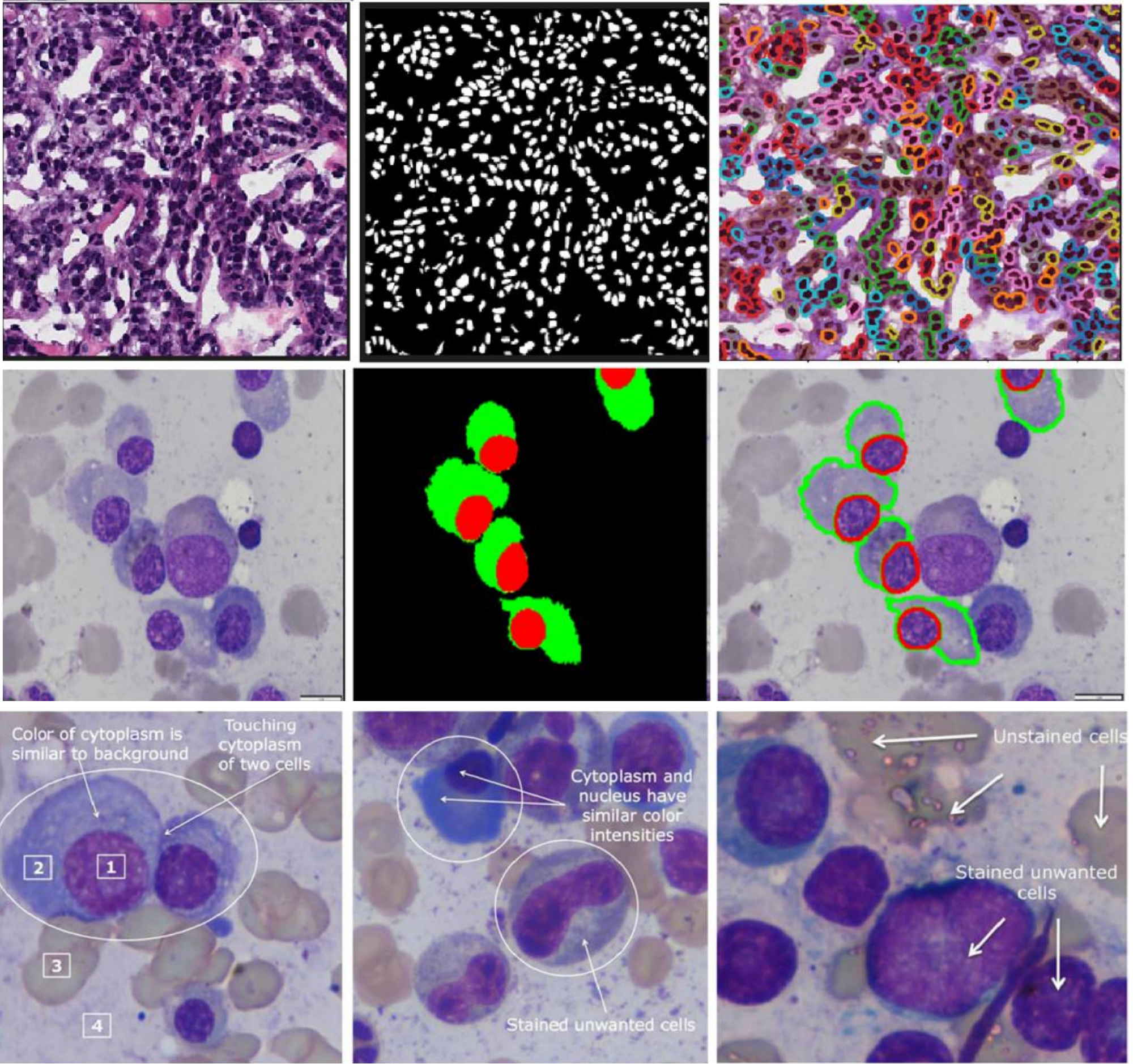}
    \caption{Challenges in microscopic image segmentation across different datasets. \textit{First row:} MoNuSeg dataset~\cite{kumar2017dataset} sample image (Column 1), ground-truth mask (Column 2), and ground-truth overlay showcasing dense, crowded, and chromatic-sparse nuclei (Column 3).  \textit{Second row:} SegPC ~\cite{gupta2021segpc} sample image (Column 1) with corresponding masks and ground-truth overlay showing clustered cells, variations in cell shape, and scale (Columns 2, 3). \textit{Third row:} Additional challenges in~SegPC.}
    \label{fig:challenge}
\end{wrapfigure}

Microscopic image segmentation assigns pixel-precise semantic labels to microscopic images and has demonstrated a wide range of applications in biological research and medical diagnosis~\cite{leygeber2019analyzing}. For instance, it enables the identification and separation of  cell regions or tissues within microscopic images, facilitating a more in-depth analysis of tissue characteristics such as size, shape, texture, and distribution~\cite{al2018fully}. Such information is crucial for understanding cellular behavior, disease diagnosis, and drug discovery. However, microscopic image segmentation is challenging due to multiple reasons, such as the presence of noise, occlusions, and overlapping cells, as shown in Figure \ref{fig:challenge}.



 Existing microscopic image segmentation approaches are generally based on   convolutional neural networks (CNNs) ~\cite{fu2018multi,vigueras2019fully}.  
 In particular, the U-Net architecture \cite{ronneberger2015u}, with its elegant U-shape encoder-decoder structure and the use of multi-scale features is effective in capturing complex image features. As a result, U-Net and its variants have become a cornerstone  for various medical image segmentation tasks, including microscopic image segmentation.  Despite the impressive performance of CNN-based methods, they often struggle to capture global dependencies due to their limited local receptive fields. The ability to capture long-range global dependencies is crucial for leveraging global information such as illumination levels, scale variations, and occlusions. To tackle this challenge, techniques such as dilated convolution~\cite{wang2020u,chen2014semantic} and channel/spatial attention models~\cite{sinha2020multi} have shown promise. Nevertheless, these techniques continue to face difficulties in achieving comprehensive global dependencies~\cite{cao2023swin, chen2021transunet}, which can potentially result in a decline in segmentation performance.


In contrast, transformers have showcased exceptional performance in capturing long-range dependencies, both in natural language processing (NLP)\cite{devlin2018bert,radford2018improving} and computer vision\cite{dosovitskiy2020image, zhang2021pyramid}, thanks to their self-attention mechanism. Nonetheless, the hierarchical nature of these transformers can restrict their capacity to learn local contextual information, as global features tend to dominate. Consequently, suboptimal segmentation outcomes may arise. To address this challenge, hybrid CNN-transformer-based methods are introduced to leverage the strengths of both types.  For instance, TransFuse~\cite{zhang2021transfuse} employs a parallel architecture that combines both CNN and transformer models, effectively merging multi-level features and resulting in an improved representation of global context. MBT-Net \cite{zhang2021multi} employs a hybrid residual transformer module that harnesses the strengths of both CNNs and transformer blocks, enabling the capture of both local details and global semantics. Additionally, it incorporates a body-edge branch to enhance local consistency. However, directly integrating local representations into self-attention transformers can be challenging, as global features may dominate local ones. This can result in reduced performance, particularly when dealing with small or indistinct objects with blurred boundaries. Consequently, developing a model that can effectively capture both local and global representations simultaneously remains a significantly challenging task.

Despite several attempts to preserve the structure and size during segmentation, accurate localization remains a challenging task. UCTransNet~\cite{wang2022uctransnet} employs multiscale channel-wise fusion with cross-attention, albeit with quadratic computational complexity, and lacks explicit consideration of local features, which are crucial for capturing scale variations, particularly in small regions. In contrast, we propose a novel approach called SA2-Net, a scale-aware attention network designed to effectively manage the diverse and intricate structures of microscopic objects, such as cells. Our primary objective is to incorporate a scale-aware module capable of effectively handling the various scales and shapes of cells or regions present in the input image.

\noindent\textbf{Contributions:} The key contributions of this paper are the following: (i) a microscopic image segmentation network known as SA2-Net, consisting of an encoder and a decoder; (ii) in the decoder, the introduction of the scale-aware attention (SA2) module, which effectively captures scale and shape variations  through local scale attention at each stage and global scale attention across scales using multi-resolution features; (iii) progressive refinement and upsampling of scale-enhanced features via the adaptive up-attention (AuA) module to generate the final segmentation mask; (iv) comprehensive experiments performed on five challenging datasets, including four microscopic segmentation datasets, demonstrate the merits of the proposed  SA2-Net.

\section{Related Work}
\label{sec:related}

\textbf{CNN-based Segmentation Networks:}
Inspired by the success of seminal U-Net \cite{ronneberger2015u}, it has
been incorporated for diverse medical image segmentation tasks. Zhou et al.~\cite{zhou2018unet++} introduced a nested U-Net (Unet++) which adds a series of nested and skip pathways in the original U-Net, enabling the capability of feature sharing between each sub-network. 
Numerous attempts have been undertaken, demonstrating encouraging results, including the utilization of dilated convolution ~\cite{wang2020u,chen2014semantic} and attention models ~\cite{sinha2020multi}. 
Moreover, a line of research has strived for the expedient design of skip-connections which is crucial for accurate segmentation as it can assist feature fusion between the expanding and contracting
paths~\cite{oktay2018attention,xu2023dcsau}.

\noindent\textbf{Transformers-based Segmentation Networks:}
Recently, transformers have shown the ability to extract global information, which is the shortcomings of CNN \cite{shamshad2023transformers}. Dosovitskiy et al.~\cite{xu2021automatic} propose the pioneering vision transformer (ViT), which achieved SOTA performance on image classification tasks by utilizing self-attention mechanisms to retain global information. In the medical domain, Swin-UNet~\cite{cao2023swin} and DS-TransUNet~\cite{lin2022ds} propose models with a U-shaped architecture that relies on the Swin Transformer \cite{liu2021swin} for 2D image segmentation. TransUNet~\cite{chen2021transunet} as a pioneer work in the direction of hybrid models, takes advantage of both CNNs and transformers to grasp both low-level and high-level features. 

\noindent\textbf{Multi-Scale Feature Learning:}
Multi-scale feature representations have lately demonstrated powerful performance in medical image segmentation due to their nature with variations in size and shape. Xu et al.~\cite{xu2021automatic} propose MSAU-Net which is a Multi-scale Attention network based on U-Net for nuclei segmentation.
CoTr~\cite{xie2021cotr} uses deformable attention to fuse flattened multi-scale feature maps from a CNN-based encoder. Gao et al.~\cite{gao2022data} proposes a multi-scale feature fusion approach that integrates spatial and semantic information globally with a linear-complexity attention mechanism. Inspired by the pyramid structure in CNNs, PVT~\cite{wang2021pyramid} presents a pyramid vision transformer. Later, Swin Transformer~\cite{liu2021swin} proposed a hierarchical vision transformer that employs shifted window for self-attention. CrossViT~\cite{chen2021crossvit}, suggests a new architecture that utilizes a dual-branch vision transformer in conjunction with a cross-attention module. 

Existing methods still struggle to localize the objects due to
unclear boundaries between tissues and surrounding areas.
In contrast, we propose SA2 module that strives to capture
explicit local information and implicit global structural information without employing self-attention to handle complex shapes and sizes of the microscopic regions such as cells. Furthermore, our A2A learns salient features of the microscopic regions  during the upsampling.
\section{Methodology}
\label{sec:method}

\noindent\textbf{Motivation:}We distinguish two important characteristics that need to be taken into account when designing an encoder-decoder framework for microscopic image segmentation. The first characteristic is multi-scale feature representation learning, which is necessary due to the significant variations in size and shape of microscopic cellular regions  across different images. To address this challenge, it is important to capture features at multiple scales that can effectively handle irregular cell shapes and sizes. The second characteristic is improved up-sampled feature discriminability for accurate localization. The lack of clear boundaries between cells and their surroundings makes it difficult to accurately segment from heavily cluttered backgrounds. To overcome this issue, an explicit attention mechanism is required during the upsampling and refinement of the features to enhance the discriminative ability for better cell localization.  Additionally, these upsampled and refined features should be enriched with high-level semantic information through deep supervision.

\begin{figure}[t]
\centering
\includegraphics[width=\textwidth]{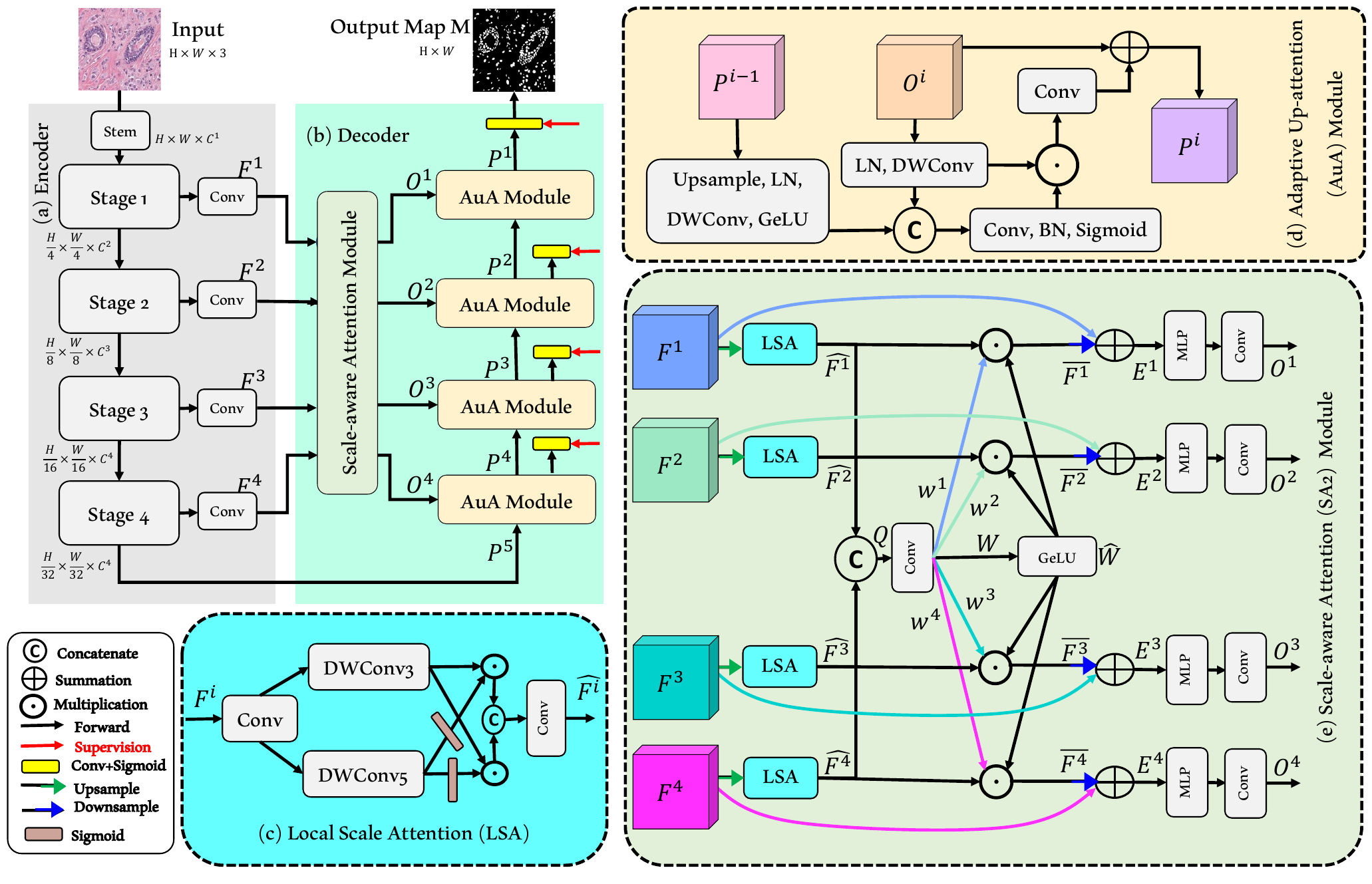}
\caption{Our proposed microscopic image segmentation network, called SA2-Net, is composed of an encoder (a) and a decoder (b). The SA2-Net takes input and uses an encoder to produce multi-resolution features at various stages. These features with different scales are then fed into a decoder, which generates the final segmentation map ($M$). The proposed decoder comprises scale-aware attention (SA2) and adaptive up-attention modules. The SA2 module (e) is designed to compute attention both for individual scales and across all scales. The proposed local scale attention (LSA) module (c) first computes attention locally for each scale. These locally attended features are then fused to produce weights for each scale ($w^i$) and global features ($W$), enabling joint capture of scale information across all scales. Finally, the adaptive up-attention-(d) module is utilized to progressively refine and upsample the scale-attended features, ultimately leading to the final mask prediction.}
\label{fig:proposed_framework}
\end{figure}

\subsection{Overall Method}
The proposed framework, named SA2-Net, is illustrated in Figure \ref{fig:proposed_framework}, which consists of an encoder that takes an input image (I) and a decoder that predicts the mask (M) for microscopic image  segmentation. The encoder generates multi-scale features for four stages with varying resolutions, which are fed into a convolution layer to generate features (F) with 64 channels, striking a balance between performance and efficiency. In the decoder, these multi-resolution features are input to the scale-aware attention (SA2) module (Fig. \ref{fig:proposed_framework}-e), which captures shape and scale variations of cells at each stage as well as jointly across all scales to produce enhanced multi-scale features for better segmentation. The enhanced multi-scale features are upsampled and refined using the adaptive up-attention (AuA) module (Fig. \ref{fig:proposed_framework}-d) to generate the prediction map (M). Additionally, deep supervision is employed at the outputs of the AuA module to accurately capture the varying shapes and sizes of foreground  regions such as cells.

\subsection{Scale-aware Attention (SA2)}
As discussed, the diverse shapes and sizes of cells necessitate the use of multi-scale features to enhance segmentation accuracy. However, we contend that directly combining these features could lead to redundancy and inconsistency, which might result in a sub-optimal segmentation task. Therefore, we present a novel scale-aware attention (SA2) module to adequately capture the size and shape discrepancies of the cells. Figure \ref{fig:proposed_framework}-(e) shows the architecture of our proposed SA2 module. The multi-resolution features from different encoder stages $F^i$ are input to the SA2 module.  Within the SA2 module, 
 a local scale attention (LSA)  is performed for each stage,  and a global scale attention is introduced across all stages. 
 
 \noindent\textbf{Local scale attention (LSA):} Our LSA  captures the scale variations at each feature resolution using different kernels, as shown in  \ref{fig:proposed_framework}-(c).  It splits the input features $F^i$ and applies an attention mechanism formulated as the element-wise product of two parallel paths of depthwise convolutional layers, one of which is activated with the Sigmoid function. These attended features are fused via concatenation and a convolution layer to generate $\hat F^i$ features. We include depthwise convolution layers in LSA to encode information from spatially neighboring pixel positions, to capture the local image structure for effective scale information learning. Furthermore, we intend to enhance these locally attended features to across scales via computing weights for each scale. We argue that each resolution feature assists in better segmentation. To do so, locally attended features $\hat F^i$ from all scales are fused and input to a convolution layer to compute single channel weights ($w^i$) for each scale, and global features. These global features are input to a GeLU  to generate global weights ($\hat W$).
\begin{flalign}
\begin{aligned}
& Q = \mathrm{Concat}(\hat F^1, \hat F^2, \hat F^3, \hat F^4) \\
& \{ w^1, w^2, w^3, w^4 \}= \mathrm{Conv}(Q) \\
& \hat W = \mathrm{GeLU}(\mathrm{Conv}(Q)) \\
\end{aligned}
\label{equation1}
\end{flalign}

 These global weights $\hat W$ are multiplied with  $\hat F^i$ to enhance  relevant features while suppressing irrelevant features based on weights across scales. Furthermore, we perform  element-wise product between these  weighted features as the following  attention formulation:   


\begin{flalign}
\begin{aligned}
& \bar F^1 = w^1 \cdot (\hat F^1 \cdot \hat W), \quad \bar F^2 = w^2 \cdot (\hat F^2 \cdot \hat W), \\
& \bar F^3 = w^3 \cdot (\hat F^3 \cdot \hat W), \quad \bar F^4 = w^4 \cdot (\hat F^4 \cdot \hat W).
\end{aligned}
\label{equation1}
\end{flalign}

These attended multi-scale features $\bar F^i$  are passed to the MLP block after a residual  connection, which are then taken as input to a convolution layer to obtain  final scale-aware attended features $O^i$ for each scale as shown in Fig. \ref{fig:proposed_framework}-(e). Here, the MLP block has LayerNorm (LN), two convolutional layers, a GeLU, and a residual depthwise convolution to incorporate the positional information as  below: 

\begin{equation} \label{mlp_equation}
\centering
F^i_{out} =  \text{Conv}(GeLU(\text{Conv}(DWConv(LN(F^i_{in})) + LN(F^i_{in})))) + F^i_{in}.
\end{equation}

\subsection{Adaptive up-attention (AuA)}
Generally, in the decoder, the features are upsampled via upsampling and convolution layers which results in suboptimal prediction masks. On the contrary, we propose an adaptive up-attention module (AuA) that upsamples and presents an attention mechanism to enrich each stage feature and strives to learn the salient features of the cells using deep supervision. 
As shown in Fig. \ref{fig:proposed_framework}-(d), AuA takes input from the previous stage features ($P^{i-1}$) and current stage scale-aware attended ($O^i$), which upsamples the previous stage features and strives to capture the salient features for the current stage via Sigmoid function. Furthermore, the outputs from each stage ($P^i$) are supervised to capture the boundaries of the cells, shown in figure \ref{fig:proposed_framework}-(b). During the inference, the prediction from the last stage $P^1$ is used to obtain the final segmentation mask $M$.

\section{Experiments}
\label{sec:experiments}

\begin{wraptable}{r}{6cm}
\centering
\scalebox{0.6}{
\begin{tabular}{|l|c|c|c|c|}
\hline
\multicolumn{1}{|c}{\multirow{2}{*}{Method}} & \multicolumn{2}{|c|}{SegPC21} & \multicolumn{2}{c|}{ISIC2018}   \\ \cline{2-5} 
\multicolumn{1}{|c}{}  & \multicolumn{1}{|l|}{Dice (\%)} & \multicolumn{1}{l|}{IoU (\%)} & \multicolumn{1}{l|}{Dice (\%)} & \multicolumn{1}{l|}{IoU (\%)} \\ \cline{2-5} 
\hline
U-Net~\cite{ronneberger2015u}      & 88.08 & 88.2 & 86.71 & 84.91 \\
UNet++~\cite{zhou2018unet++}    & 91.02 & 90.92 & 88.22 & 86.51 \\
AttUNet~\cite{oktay2018attention}    & 91.58 & 91.44 & 88.20 & 86.49 \\
MultiResUNet~\cite{ibtehaz2020multiresunet}    & 86.49 & 86.76 & 86.94 & 85.37 \\
TransUNet~\cite{chen2021transunet}  & 82.33 & 83.38 & 84.99 & 83.65 \\
ResidualUNet~\cite{zhang2018road}      & 84.79 & 85.41 & 86.89 & 85.09 \\
MissFormer\cite{huang2021missformer}  & 80.82 & 82.09 & 86.57 & 84.84 \\
UCTransNet~\cite{wang2022uctransnet} & 91.74 & 91.59 & \textbf{88.98} & \textbf{87.29} \\
\hline
\textbf{SA2-Net (Ours)}   & \textbf{92.41}  & \textbf{92.23}   & {88.88}  & {87.21} \\
\hline
\end{tabular}}
\caption{Comparison of our method over SegPC21 and ISIC2018 dataset with SOTA methods using Dice and IoU metrics. The best results are in bold.}
\label{tab:segpc-isic}
\end{wraptable}

\subsection{Datasets} We perform  experiments on five  medical/microscopic image segmentation benchmark datasets, including  MoNuSeg~\cite{kumar2017dataset}, SegPC-2021~\cite{gupta2021segpc}, GlaS~\cite{sirinukunwattana2017gland},  ISIC-2018~\cite{codella2019skin}, and ACDC \cite{chen2021transunet}.  
The Multiorgan Nucleus Segmentation (MoNuSeg) dataset~\cite{kumar2017dataset} has been generated utilizing H\&E-stained cell images. The training set comprises 30 high-resolution images, which have undergone manual annotation with approximately 22,000 nuclear boundary annotations. In comparison, the test set contains 14 images labelled with over 7000 nuclear boundary annotations.
Our methodology is also evaluated on the SegPC2021 dataset~\cite{gupta2021segpc} for the segmentation of multiple myeloma cells. The dataset consists of a training set comprising 290 samples, and two other sets used for validation and testing containing 200 and 277 samples respectively. To evaluate our method on this dataset, we follow the same strategy presented in \cite{azad2022medical}. 
The Gland Segmentation in Colon Histology Images (GlaS)~\cite{sirinukunwattana2017gland} dataset includes high-resolution microscopic images of Hematoxylin and Eosin (H\&E) stained slides and corresponding ground truth annotations which comprise a total of 165 images, including 85 images for training and 80 images for testing purposes.
Moreover, we used the ISIC-2018 dataset~\cite{codella2019skin}, collected by the International Skin Imaging Collaboration (ISIC) as a large-scale dataset of 2594 dermoscopy images along with their corresponding ground truth annotations and followed \cite{asadi2020multi} in splitting and pre-processing the data. 
Lastly, we used the ACDC dataset containing 100 MRI scans, where multiple patients were scanned, and each scan was labelled for three different organs which are the left ventricle (LV), right ventricle (RV), and myocardium (MYO). The training, validation, and testing process follow literature work \cite{cao2023swin}, where 70 cases are used for training, 10 for validation, and 20 for testing purposes.

\begin{table}[t]
\centering
\scalebox{0.7}{
\begin{tabular}{|l|c|c|c|c|}
\hline
\multicolumn{1}{|c}{\multirow{2}{*}{Method}} & \multicolumn{2}{|c|}{GlaS} & \multicolumn{2}{c|}{MoNuSeg}  \\ \cline{2-5} 
\multicolumn{1}{|c}{}  & \multicolumn{1}{|l|}{Dice (\%)} & \multicolumn{1}{l|}{IoU (\%)} & \multicolumn{1}{l|}{Dice (\%)} & \multicolumn{1}{l|}{IoU (\%)} \\ \cline{2-5} 
\hline
U-Net~\cite{ronneberger2015u}      & 85.45 $\pm$ 1.3 & 74.78 $\pm$ 1.7 & 76.45 $\pm$ 2.6 & 62.86 $\pm$ 3.0 \\
UNet++~\cite{zhou2018unet++}      & 87.56 $\pm$ 1.2  & 79.13  $\pm$ 1.7  & 77.01 $\pm$ 2.1 & 63.04 $\pm$ 2.5 \\
AttUNet~\cite{oktay2018attention}     & 88.80 $\pm$ 1.1   & 80.69 $\pm$ 1.7  & 76.67 $\pm$ 1.1 & 63.47 $\pm$ 1.2 \\
MultiResUNet~\cite{ibtehaz2020multiresunet}  & 88.73 $\pm$ 1.2  & 80.89 $\pm$ 1.7  & 78.22 $\pm$ 2.5 & 64.83$\pm$ 2.9 \\
TransUNet~\cite{chen2021transunet}   & 88.40 $\pm$ 0.7  & 80.40$\pm$ 1.0  & 78.53 $\pm$ 1.1 & 65.05 $\pm$ 1.3 \\
MedT~\cite{valanarasu2021medical}        & 85.93 $\pm$ 2.9  & 75.47 $\pm$ 3.5  & 77.46 $\pm$ 2.4 & 63.37 $\pm$ 3.1 \\
Swin-Unet~\cite{cao2023swin}   & 89.58 $\pm$ 0.6  & 82.07 $\pm$ 0.7  & 77.69 $\pm$ 0.9 & 63.77 $\pm$ 1.2 \\
UCTransNet~\cite{wang2022uctransnet}  & 90.18 $\pm$ 0.7  & 82.96 $\pm$ 1.1  & 79.08 $\pm$ 0.7 & 65.50 $\pm$ 0.9  \\
\hline
\textbf{ SA2-Net (Ours)}   & \textbf{91.38 $\pm$ 0.4}  & \textbf{84.90 $\pm$ 0.6}   & \textbf{81.34 $\pm$ 0.5}  & \textbf{68.70 $\pm$ 0.7}                      \\
\hline
\end{tabular}}
\caption{Comparison of our method over GlaS and MoNuSeg dataset with SOTA methods using Dice and IoU metrics. The best results are in bold.}
\label{tab:glasMonuseg}
\end{table}



\begin{wraptable}{r}{6cm}
\vspace{-1cm}
    \centering
    \scalebox{0.7}{
\begin{tabular}{lcccc}
\hline
{Method} & {Dice Avg} & {RV} & {Myo} & {LV} \\ \hline
U-Net \cite{ronneberger2015u} & 87.55 & 87.10 & 80.63 & 94.92 \\
R50+AttnUNet \cite{chen2021transunet} & 86.75 & 87.58 & 79.20 & 93.47 \\
ViT+CUP \cite{chen2021transunet} & 81.45 & 81.46 & 70.71 & 92.18 \\
R50+ViT+CUP \cite{chen2021transunet} & 87.57 & 86.07 & 81.88 & 94.75 \\
TransUNet \cite{chen2021transunet} & 89.71 & 86.67 & 87.27 & 95.18 \\
SwinUnet \cite{cao2023swin} & 90.00 & 88.55 & 85.62 & {95.83} \\ \hline
\textbf{SA2Net (Ours)} & \textbf{92.30} & \textbf{90.33} & \textbf{90.39} & \textbf{96.18} \\ \hline
\end{tabular}}
\caption{Comparison of our approach over
ACDC dataset reported an average Dice score metric and Dice score for each organ. The best results are in bold.}
\label{tbl:comparision_acdc}
\end{wraptable}
\vspace{-0.1cm}
\subsection{Implementation Details}
\label{sec:implementation}
Our model was developed using PyTorch on a single NVIDIA RTX A6000 GPU card with 48 GB memory. The input resolution was set to $224\times224$, and we opted not to use pre-trained weights for training. During training, random flipping and rotation were employed for data augmentation, and the combined weighted IoU and weighted BCE loss function was used.
Similar to UCTransNet \cite{wang2022uctransnet}, for GlaS and MoNuSeg datasets, a batch size of 4 was set, and the Adam optimizer was used with an initial learning rate of 0.001. To ensure the reliability of the results on small datasets, we conducted three times 5-fold cross-validation, totaling 15 cross-validations. For the GlaS and MoNuSeg datasets, an ensemble technique was employed by obtaining predictions from all five models and taking the mean of these predictions as the final prediction.
Similar to \cite{azad2022medical}, for SegPC21 and ISIC2018, a batch size of 16 was set, and training was done for 100 epochs using the Adam optimizer with an initial learning rate of 0.0001. For ACDC, a batch size of 12 was set, and training was done for 150 epochs using the Adam optimizer with an initial learning rate of 0.0001. Sigmoid activation was applied for the binary segmentation mask and Softmax activation for multi-class segmentation. Our method was evaluated using dice and IoU metrics.

\begin{figure*}[t]
    \centering
    \resizebox{0.65\textwidth}{!}{
    \begin{tabular}{@{} *{6}c @{}}
    \includegraphics[width=0.28\textwidth]{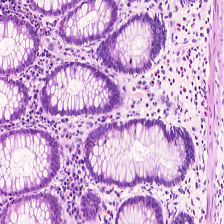} &
    \includegraphics[width=0.28\textwidth]{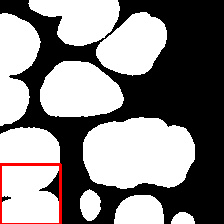} &
    \includegraphics[width=0.28\textwidth]{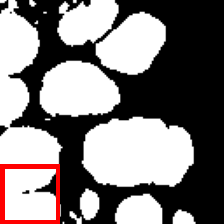} &
    \includegraphics[width=0.28\textwidth]{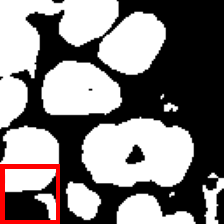} &
    \includegraphics[width=0.28\textwidth]{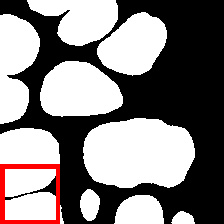} &
    \includegraphics[width=0.28\textwidth]{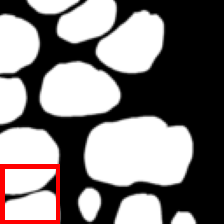} \\
    \includegraphics[width=0.28\textwidth]{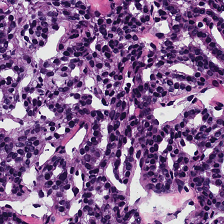} &
    \includegraphics[width=0.28\textwidth]{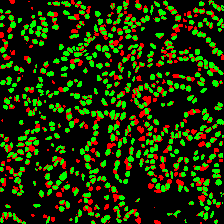} &
    \includegraphics[width=0.28\textwidth]{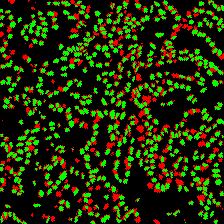} &
    \includegraphics[width=0.28\textwidth]{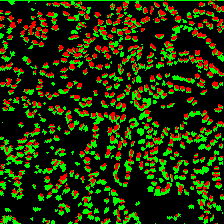} &
    \includegraphics[width=0.28\textwidth]{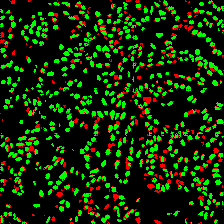} &
    \includegraphics[width=0.28\textwidth]{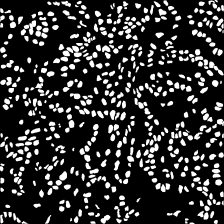} \\
    { \large (a) Input Image} & {\large (b) UCTransNet} & {\large (c) MedT} & {\large (d) UNet} & {\large (e) SA2-Net} & {\large (f) Ground truth}
    \end{tabular}
}
    \caption{Qualitative results on Glas and MoNuSeg datasets. The \textcolor{red}{red box}  in the top row highlights regions where exactly SA2-Net performs better than the other methods, while in the bottom row \textcolor{green}{green} indicates true positive and \textcolor{red}{red} shows prediction errors.} 
    \label{fig:glasmono}
    \vspace{-1em}
\end{figure*}

\begin{figure*}[t]
    \centering
    \resizebox{0.65\textwidth}{!}{
    \begin{tabular}{@{} *{6}c @{}}
    \includegraphics[width=0.28\textwidth]{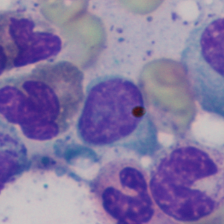} &
    \includegraphics[width=0.28\textwidth]{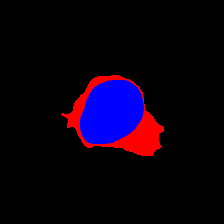} &
    \includegraphics[width=0.28\textwidth]{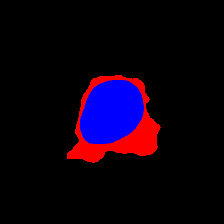} &
    \includegraphics[width=0.28\textwidth]{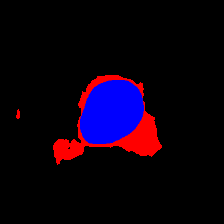} &
    \includegraphics[width=0.28\textwidth]{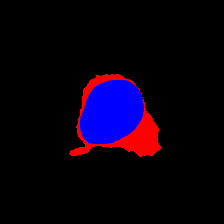} &
    \includegraphics[width=0.28\textwidth]{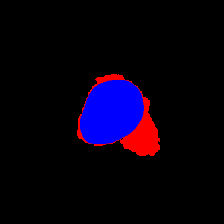} \\
    \includegraphics[width=0.28\textwidth]{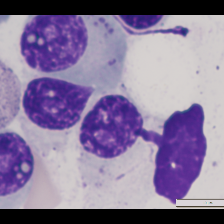} &
    \includegraphics[width=0.28\textwidth]{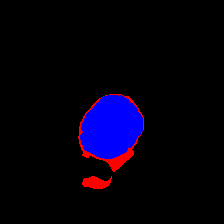} &
    \includegraphics[width=0.28\textwidth]{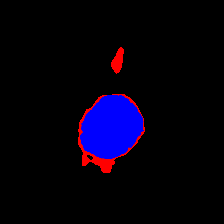} &
    \includegraphics[width=0.28\textwidth]{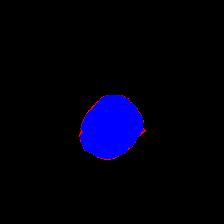} &
    \includegraphics[width=0.28\textwidth]{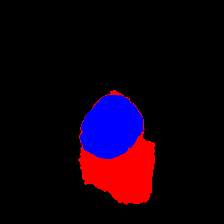} &
    \includegraphics[width=0.28\textwidth]{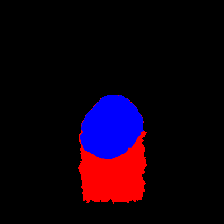} \\
    { \large (a) Input Image} & {\large (b) U-Net} & {\large (c) TransUNet} & {\large (d) UCTransNet} & {\large (e) SA2-Net} & {\large (f) Ground truth}
    \end{tabular}
}
    \caption{Qualitative comparison of the proposed method on the \textit{SegPC-2021} dataset for segmentation of Multiple Myeloma plasma cells. For a fair comparison, we follow \cite{azad2022medical} and  perform the cytoplasm segmentation (\textcolor{red} {red mask}) for a given input  nucleus  mask (\textcolor{blue}{blue}).} 
    \label{fig:segpc}
    \vspace{-1em}
\end{figure*}

\begin{figure}[!t]
\centering
\includegraphics[width=0.65\textwidth]{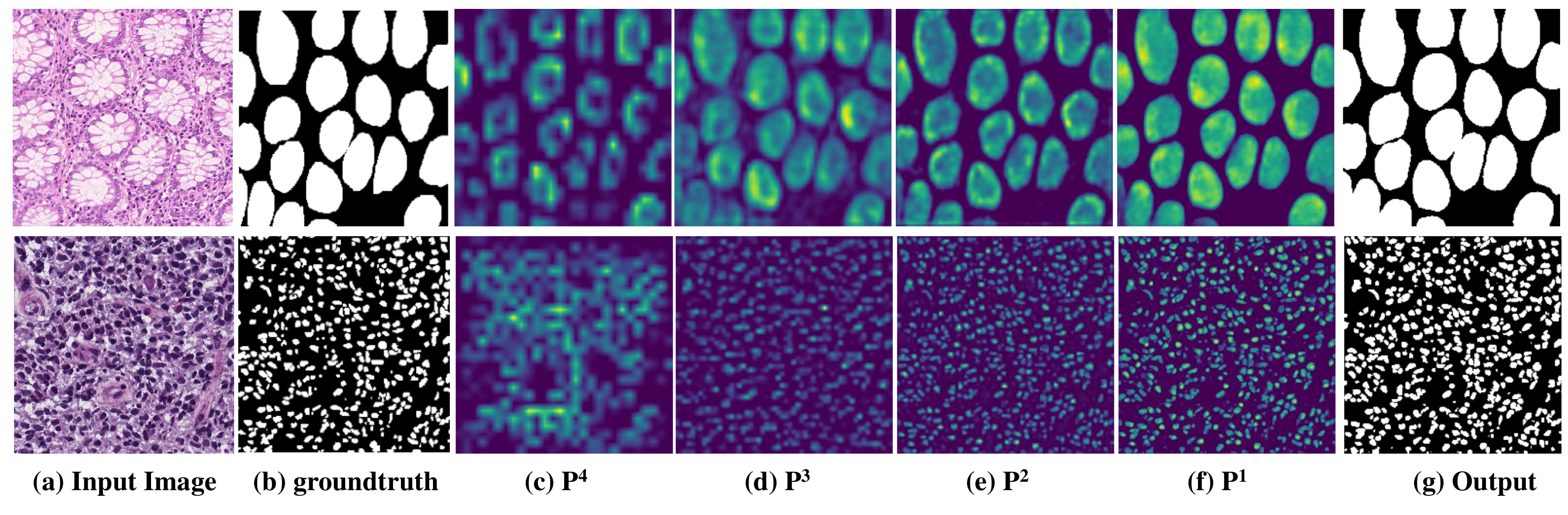}
\caption{Feature visualization of our the AuA modules output ($\textit{P}^{4}$ to $\textit{P}^{1}$) using Grad-CAM~\cite{selvaraju2017grad} over GlaS (first row) and MoNuSeg (second row) images. The results show that with the help of the SA2 module, AuA modules, and deep supervision,  features that are robust to the shape, size, and density variations of cells are learned at the decoder. }
 \label{fig:gradcam}
  \vspace{-1em}
\end{figure}
\subsection{Quantitative Results}
The comparison of the proposal with previous SOTA methods is presented in Table \ref{tab:glasMonuseg}, \ref{tab:segpc-isic}, and \ref{tbl:comparision_acdc} respectively. Table \ref{tab:glasMonuseg} shows that the proposed SA2-Net has significant improvements over prior arts for GlaS and MoNuSeg datasets, e.g., our method gains a dice score of 1.20\% and 2.67\% over best-performing UCTransNet \cite{wang2022uctransnet}. A similar trend can be also witnessed using IoU metric, where our method outperforms SOTA over GlaS and MoNuSeg by a large margin and achieves 84.90 and 68.70 scores respectively. Table \ref{tab:segpc-isic} presents the results for SegPC and ISIC datasets, which highlights that our method significantly outperforms other methods for SegPC and obtains comparable results compared to UCTransNet \cite{wang2022uctransnet} over ISIC2018 dataset. Also, our method achieves SOTA over the ACDC dataset shown in table \ref{tbl:comparision_acdc}.

\subsection{Qualitative Results}

We also provide qualitative comparison results on three datasets to further prove the generalization ability of our SA2-Net as shown in Figures \ref{fig:glasmono} and \ref{fig:segpc}. 
Looking at Figure \ref{fig:glasmono}, our predictions on Glas and MoNuSeg datasets adjust well to the provided GT masks indicating that our method has the capability to create multi-scale representation and can filter out background noise which is common in datasets that have overlapping backgrounds.
Furthermore, the SegPC dataset's image outputs (Figure \ref{fig:segpc}) illustrate that SA2-Net has the ability to grasp intricate details, create precise contours, and exhibit strong performance in areas where boundaries can be unclear due to interference, showing robustness to noisy items adapting to the given GT masks. This can be attributed to the expedient combination of attention-based modules and CNN for modeling global connections and local representations.


\begin{wraptable}{r}{6cm}
    \centering
    \scalebox{0.5}{
    \begin{tabular}{cccccccc}
    \hline
     \textbf{Model} & \textbf{SA2} & \textbf{LSA}  & \textbf{AuA} & \textbf{Deep Supervision} & \textbf{Dice} & \textbf{IoU}
     \\ \hline
     Baseline (UNet)   & \xmark   & \xmark    & \xmark & \xmark & 76.55 & 62.76    \\
    SA2-Net   & \checkmark    & \xmark   & \xmark & \xmark & 80.25 & 68.96    \\
    SA2-Net   & \xmark     & \checkmark  & \xmark & \xmark & 79.98 & 68.84    \\
    SA2-Net   & \checkmark & \checkmark  & \xmark  & \xmark & 81.53 & 69.12   \\
    SA2-Net   & \checkmark & \checkmark  & \checkmark  & \xmark & 81.65 & 69.20   \\
    SA2-Net   & \checkmark & \checkmark & \checkmark & \checkmark & \textbf{81.75} & \textbf{69.25}      \\ \hline
    \end{tabular}}
        \caption{
       Ablation experiments on MoNuSeg dataset:  impact of individual contributions on segmentation performance of SA2-Net.}
    \label{table:ablation}    
\end{wraptable}
\vspace{-0.5cm}
\subsection{Ablation Study}\label{sec:ablation}
We performed thorough  ablation study further to validate the performance of each module in SA2-Net under different settings.  Specifically, we investigate the influence of each contribution and the results are summarized in Table \ref{table:ablation}. The combination of SA2 and LSA in SA2-Net shows that the two attention mechanisms are complementary and can collaborate to provide better segmentation predictions. However, an attention-based signal for our decoder using deep supervision produces the best performance in all metrics as it captures local contextual relations among pixels.  Lastly, we present the feature visualization using Grad-CAM~\cite{selvaraju2017grad}  of each level of our AuA module (Figure \ref{fig:gradcam}). As illustrated, after applying the attention mechanism at each level, more emphasis is drawn to the desired region and is more highlighted than the surroundings, demonstrating that the SA2 module, AuA modules, and deep supervision learn features that are robust to the shape, size, and density variations of cells. Furthermore, each level serves a complimentary function, with the larger level providing fine-grained features and the smaller level attempting to give extra information. As a result, all levels are required for the model to function effectively.
\vspace{-0.5cm}
\section{Conclusion}
\label{sec:conclusion}
In this paper, we present SA2-Net, a novel framework for microscopic image segmentation.  
The proposed approach effectively captures variations in the size and shape of microscopic regions such as cells through multi-scale feature learning.   Furthermore, we incorporate local scale attention at each stage and global scale attention across scales to address the challenges posed by diverse cell structures.  Overall, the proposed framework leverages multi-scale features, scale-aware attention, adaptive up-attention, and deep supervision for accurate microscopic image segmentation.  Experiments across five diverse datasets show the superiority of the proposed method.
\vspace{-0.1cm}
\section*{Acknowledgement}
This work is partially supported by the MBZUAI-WIS Joint Program for AI Research (Project grant number- WIS P008).

\bibliography{egbib}
\end{document}